\documentclass[letterpaper]{article} 

\usepackage{aaai23}  

\usepackage{times}  
\usepackage{helvet}  
\usepackage{courier}  
\usepackage[hyphens]{url}  
\usepackage{graphicx} 
\usepackage{natbib}  
\usepackage{caption} 
\usepackage{algorithm}
\usepackage{algorithmic}
\usepackage{tablefootnote}

\usepackage{newfloat}
\usepackage{listings}
\DeclareCaptionStyle{ruled}{labelfont=normalfont,labelsep=colon,strut=off} 
\floatstyle{ruled}
\newfloat{listing}{tb}{lst}{}
\floatname{listing}{Listing}
\usepackage{algorithm}
\usepackage{algorithmic}

\usepackage{url}
\usepackage{amsmath}
\usepackage{caption}
\usepackage{subcaption}
\usepackage{amsmath,amsfonts,amssymb}
\usepackage{mathrsfs}
\usepackage{multirow}
\usepackage{makecell}
\usepackage{colortbl}
\usepackage{booktabs}
\usepackage{array}
\usepackage{newfloat}
\usepackage{listings}
\definecolor{lightgray}{rgb}{0.9, 0.9, 0.9}
\definecolor{ballblue}{rgb}{0.13, 0.67, 0.8}
\newcommand{\STAB}[1]{\begin{tabular}{@{}c@{}}#1\end{tabular}}

\title{FacT: Factor-Tuning for Lightweight Adaptation on Vision Transformer}

\author {
    Shibo Jie,
    Zhi-Hong Deng\thanks{Corresponding Author.}
}
\affiliations {
    School of Intelligence Science and Technology, Peking University\\

    \{parsley, zhdeng\}@pku.edu.cn
}

\usepackage{bibentry}

\begin{document}

\maketitle

\begin{abstract}
Recent work has explored the potential to adapt a pre-trained \emph{vision transformer} (ViT) by updating only a few parameters so as to improve storage efficiency, called \emph{parameter-efficient transfer learning} (PETL). Current PETL methods have shown that by tuning only 0.5\% of the parameters, ViT can be adapted to downstream tasks with even better performance than full fine-tuning. In this paper, we aim to further promote the efficiency of PETL to meet the extreme storage constraint in real-world applications. To this end, we propose a tensorization-decomposition framework to store the weight increments, in which the weights of each ViT are tensorized into a single 3D tensor, and their increments are then decomposed into lightweight factors. In the fine-tuning process, only the factors need to be updated and stored, termed \emph{Factor-Tuning} (\texttt{\textbf{FacT}}). On VTAB-1K benchmark, our method performs on par with NOAH, the state-of-the-art PETL method, while being 5$\times$ more parameter-efficient. We also present a tiny version that only uses 8K (0.01\% of ViT's parameters) trainable parameters but outperforms full fine-tuning and many other PETL methods such as VPT and BitFit. In few-shot settings, \texttt{\textbf{FacT}} also beats all PETL baselines using the fewest parameters, demonstrating its strong capability in the low-data regime. 
\end{abstract}

\section{Introduction}

The de-facto paradigm for achieving state-of-the-art performance on visual tasks involves pre-training on large datasets like ImageNet~\cite{imagenet}, and then fully fine-tuning on downstream tasks. However, this paradigm is not storage-efficient because each downstream task necessitates the storage of an entire model, which becomes prohibitive in some scenarios (\emph{e.g.}, storing customized models for each user) as the size of vision models grows exponentially. 

To promote storage efficiency, recent work on \emph{parameter-efficient transfer learning} (PETL) attempts to fine-tune only a small part of the parameters to adapt the large pre-trained models to downstream tasks. These methods either insert additional trainable structures (\emph{e.g.}, adapters~\cite{adapter} or prompt tokens~\cite{vpt}) to a frozen \emph{vision transformer} (ViT)~\cite{vit} backbone or selectively fine-tune some of the ViT's own parameters (\emph{e.g.}, all bias parameters~\cite{bitfit}). Recently, LoRA~\cite{lora} also shows that optimizing the low-rank decomposition matrices of \emph{weight increments}\footnote{We refer to the weight's {change} during fine-tuning as its \emph{increment}, which is a matrix or tensor of the same shape as the weight.} of the dense layers is a promising way to adapt large models.

However, though LoRA's matrix decomposition significantly reduces the storage overhead of fine-tuned dense layers, it is far from exploiting the low-rank properties of neural networks to the extreme. Inspired by recent work on compression of the transformer-based \emph{pre-trained language models} (PLMs)~\cite{inter2}, we infer that during fine-tuning, the weight increments of pre-trained ViT are also redundant in terms of \emph{intra-weight rank} and \emph{inter-weight rank}. The former is reflected in that the dense increments matrix can be low-rank as in LoRA, while the latter is implied by the fact that cross-layer weight-sharing has already been adopted in some lightweight ViT structures~\cite{minivit}, which is not taken into consideration by LoRA. 
\begin{figure}[t]
     \centering
         \includegraphics[width=0.40\textwidth]{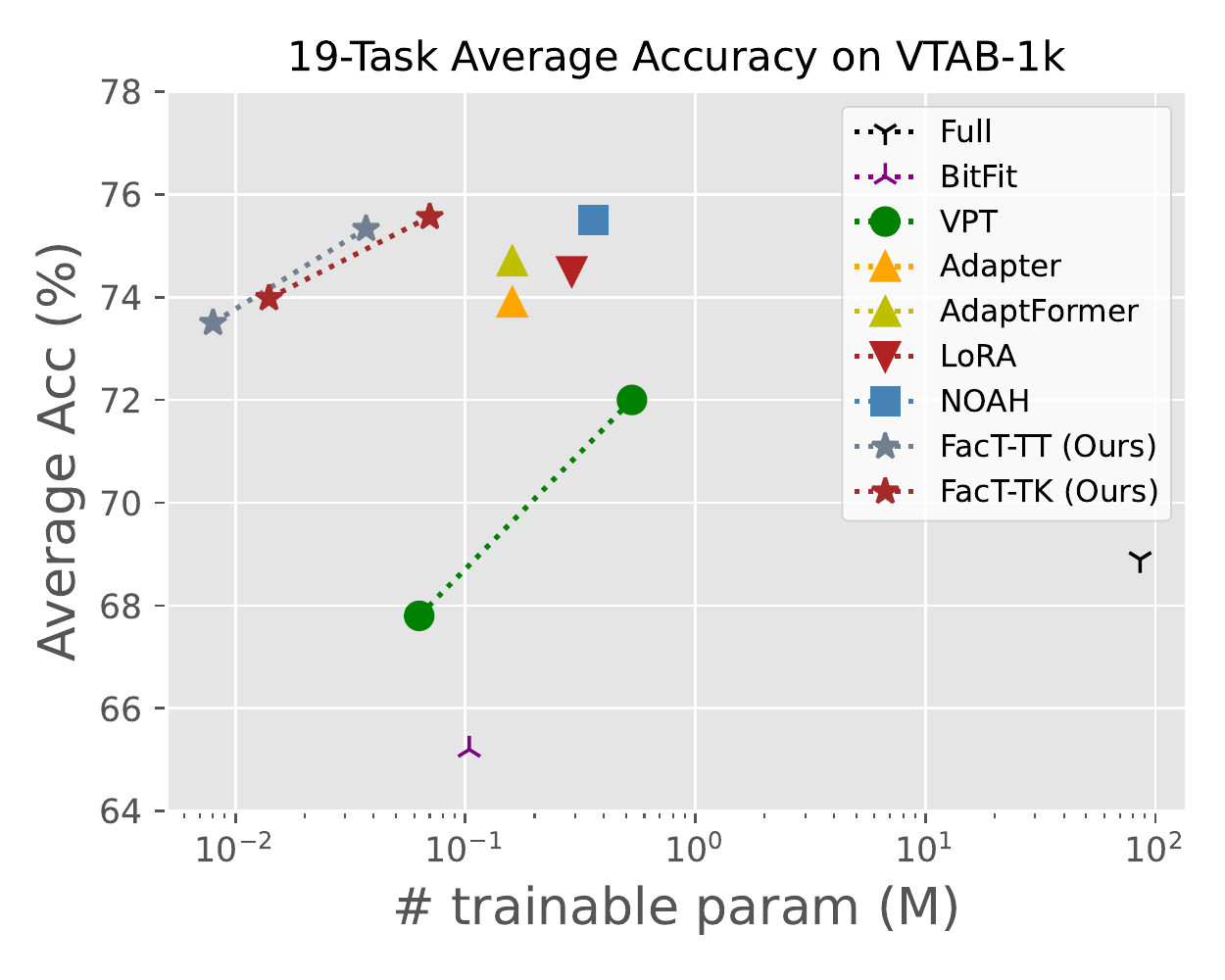}
         \caption{{Average accuracy \emph{vs.} number of trainable parameters (log axis) on VTAB-1K benchmark.} Our \texttt{\textbf{FacT}} significantly reduces the number of trainable parameters.}
         \label{fig:intro}
\end{figure}
To fully develop the potential of PETL, we propose a \emph{tensorization-decomposition} framework. Instead of decomposing the weight increment matrices individually like LoRA, we represent the whole ViT as a single tensor, in which all weight increment matrices are decomposed together. This means that the different matrices could share their factors to reduce inter-weight rank redundancy as well. The factors are then fine-tuned with the ViT backbone frozen, called \emph{Factor-Tuning} (\texttt{\textbf{FacT}}). After fine-tuning, only the classification head and these factors need to be stored, and thus the storage efficiency is significantly promoted.

In this paper, we explore the feasibility of applying various tensor decomposition formats to the tensorization-decomposition framework, including \emph{Tensor-Train} format~\cite{tt} (denoted as \texttt{\textbf{FacT-TT}}) and \emph{Tucker} format~\cite{tucker} (denoted as \texttt{\textbf{FacT-TK}}). We also show that LoRA can be regarded as a special case of \texttt{\textbf{FacT}} when using \emph{Matrix-Batch} format. As the experimental results on VTAB-1K benchmark shown in Fig~\ref{fig:intro}, both \texttt{\textbf{FacT-TT}} and \texttt{\textbf{FacT-TK}} significantly reduce the number of trainable parameters while maintaining competitive performance. \texttt{\textbf{FacT-TK}} achieves performance on par with that of the current state-of-the-art (SOTA) method NOAH~\cite{noah} with only 19\% of NOAH's parameters. In a more extreme case, \texttt{\textbf{FacT-TT}} can adapt the ViT with 85.8M parameters by only tuning the factors with 8K parameters, while still outperforming full fine-tuning and some other PETL methods such as VPT~\cite{vpt} and BitFit~\cite{bitfit}. In few-shot learning, \texttt{\textbf{FacT-TT}} beats all other PETL baselines, demonstrating its superiority in the low-data regime. 

The contributions of our work are as follows:
\begin{itemize}
    \item Inspired by the assumption that weight increments of pre-trained ViT during fine-tuning are rank-redundant, 
    we propose \texttt{\textbf{FacT}}, a tensorization-decomposition framework to adapt ViT by tuning the factors of increments.
    \item Under this framework, we propose \texttt{\textbf{FacT-TT}} and \texttt{\textbf{FacT-TK}}, decomposing the tensorized ViT with Tensor-Train and Tucker format, respectively.
    \item Experimental results show that our methods are much more lightweight than other PETL methods yet maintain competitive performances on VTAB-1K benchmark and SOTA results on few-shot learning datasets, demonstrating the potential of PETL with an extremely small storage budget.
\end{itemize}

\section{Related Work}
\subsection{Parameter-Efficient Transfer Learning}
PETL has been investigated in the field of both \emph{natural language processing} (NLP) and \emph{computer vision}, which aims to fine-tune a small number of trainable parameters to transfer large pre-trained models to downstream tasks. We here introduce some PETL methods either proposed for ViT or ported from PLMs. 

\textbf{Adapter}~\cite{adapter,compactor,adapter-cv} is typically a bottleneck block composed of two fully connected layers, whose weights are $\boldsymbol{W}_{down}\in \mathbb{R}^{d\times h}$ and $\boldsymbol{W}_{up}\in \mathbb{R}^{h\times d}$, where $h << d$. There are two common ways to insert adapters. The original way is sequential~\cite{adapter,adapterp}, formulated as
$$\boldsymbol{X}'\leftarrow\boldsymbol{X}+\phi(\boldsymbol{X}\boldsymbol{W}_{down})\boldsymbol{W}_{up}$$
where $\boldsymbol{X}\in \mathbb{R}^{N\times d}$ is the output of Feed-Forward Network (FFN) blocks and $\phi$ is a nonlinear function. The other is parallel~\cite{towards,adaptformer}, formulated as $$\boldsymbol{X}'\leftarrow\boldsymbol{X}+\textit{FFN}(\boldsymbol{X})+s\cdot\phi(\boldsymbol{X}\boldsymbol{W}_{down})\boldsymbol{W}_{up}$$
where $s$ is a hyper-parameter, $\boldsymbol{X}$ is the input of FFN blocks. The parallel adapter design is also referred to as \textbf{AdaptFormer} by \citet{adaptformer}. Concurrently, \citet{convpass} suggest using convolutional adapters for ViT. \citet{ssf} propose shifting and scaling the intermediate features of the models which can also be regarded as a simplified linear adapter.

\textbf{LoRA}~\cite{lora} decomposes the increments of query transformation $\boldsymbol{W}_{q}$ and value transformation $\boldsymbol{W}_{v}$ into low-rank $\boldsymbol{A}_{q/v}\in \mathbb{R}^{d\times r}$ and $\boldsymbol{B}_{q/v}\in \mathbb{R}^{r\times d}$ where $r << d$. The query and value are then computed as 
$$\boldsymbol{Q/V}\leftarrow\boldsymbol{XW}_{q/v}+s\cdot\boldsymbol{XA}_{q/v}\boldsymbol{B}_{q/v}$$
in which $s$ is a hyper-parameter.

\textbf{VPT}~\cite{vpt} concatenates the input $\boldsymbol{X}$ with several trainable prompts $\boldsymbol{P}\in \mathbb{R}^{l\times d}$ before feeding it into transformer layers. This extended sequence is formulated as
$$\boldsymbol{X}'\leftarrow[\boldsymbol{X}, \boldsymbol{P}]$$
In VPT-Deep, these prompts are concatenated before every layer and then discarded at the end of the layer. While in VPT-Shallow, the prompts are only inserted before the first layer and will be maintained until the last layer.

\textbf{NOAH}~\cite{noah} focuses on combining existing PETL methods without manual design. It trains a large supernet at first and then performs neural architecture search on hidden dimension $h$ of Adapter, rank $r$ of LoRA, and prompt length $l$ of VPT.

In all the aforementioned methods, the parameters of pre-trained ViT are frozen. There are also methods that only fine-tune a few of the pre-trained parameters without introducing new parameters, such as \textbf{BitFit}~\cite{bitfit}, which fine-tunes the bias parameters only. 

\subsection{Tensor Decomposition for Network Compression}
Tensor decomposition is an important research area aiming to approximate a tensor with a set of low-rank factors that has been studied for many years. Previous work on deep learning has investigated the use of tensor decomposition to compress neural networks so as to reduce the size of models, including ConvNets~\cite{com_cnn1,com_cnn2}, RNNs~\cite{com_rnn1,com_rnn2,com_rnn3}, and Transformers~\cite{intra1,inter1}.

Note that model compression and PETL have a similar purpose: to reduce the storage overhead. But the difference lies in that model compression aims at reducing the size of the whole model, while PETL only considers reducing the trainable parameters on a pre-trained model since the pre-trained weights are always required when fine-tuning on subsequent new tasks. Therefore, in this paper, we take inspiration from model compression but compress the increments of the weights instead of the pre-trained weights themselves.

\section{Method}
\begin{figure}[t]
     \centering
         \includegraphics[width=0.45\textwidth]{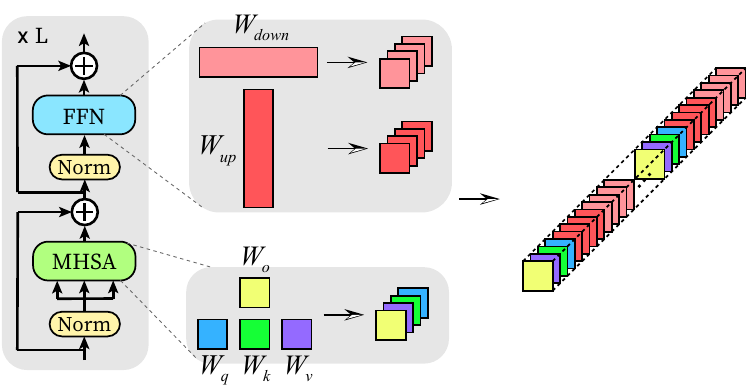}
         \caption{{Illustration of tensorizing ViT.} The ViT is tensorized into a single $12L\times d\times d$ tensor.}
         \label{fig:tensorize}
\end{figure}
\subsection{Tensorizing Vision Transformer}
Tensorizing a neural network means representing its parameters using a single tensor. Previous vision models such as ResNet~\cite{resnet} usually use weights of different sizes in different layers, \emph{e.g.}, different kernel sizes and input/output channels. This property limits their capability of tensorization. However, due to the consistency of Transformer layers in ViT, we can tensorize ViT in a much simpler way. 

Besides the patch embedding and classification head, a ViT is composed of two types of blocks: Multi-Head Self-Attention (MHSA) and Feed-Forward Network (FFN, also referred to as Multi-Layer Perceptron). In MHSA, the query, key, value, and output transformations are parametrized by $\boldsymbol{W}_q, \boldsymbol{W}_k, \boldsymbol{W}_v, \boldsymbol{W}_o \in \mathbb{R}^{d\times d}$, respectively. These transformations are further divided into $N_h$ heads: $\{\boldsymbol{W}_q^{(i)}\}^{N_h}_{i=1}, \{\boldsymbol{W}_k^{(i)}\}^{N_h}_{i=1}, \{\boldsymbol{W}_v^{(i)}\}^{N_h}_{i=1}, \{\boldsymbol{W}_o^{(i)}\}^{N_h}_{i=1}$. Then, the MHSA is formulated as\footnote{We use superscript $(i)$ to denote the $i$-th head.}
\begin{equation}
\begin{aligned}
    \textit{MHSA}(\boldsymbol{X})=&\\\sum_{i=1}^{N_h}\textit{softmax}&\left(\frac{\boldsymbol{X}\boldsymbol{W}_q^{(i)}{\boldsymbol{W}_k^{(i)}}^\intercal\boldsymbol{X}^\intercal}{\sqrt{d}}\right)\boldsymbol{X}\boldsymbol{W}_v^{(i)}{\boldsymbol{W}_o^{(i)}}^\intercal
\end{aligned}    
\end{equation}

An FFN block consists of two fully-connected (FC) layers. Ignoring the bias parameters for simplicity, the FFN is formulated as
\begin{equation}
\begin{aligned}
    \textit{FFN}(\boldsymbol{X})=\textit{GELU}(\boldsymbol{X}\boldsymbol{W}_{up})\boldsymbol{W}_{down}
\end{aligned}    
\end{equation}
where $\boldsymbol{W}_{up}\in \mathbb{R}^{d\times 4d}$ and $\boldsymbol{W}_{down}\in \mathbb{R}^{4d\times d}$ are the weights of the FC layers.

The FFN can also be regarded as a multi-head block. We divide $\boldsymbol{W}_{up}$ and $\boldsymbol{W}_{down}$ into four matrices of size $d\times d$, \emph{i.e.}, $\{\boldsymbol{W}_{up}^{(i)}\}^{4}_{i=1}$ and $\{\boldsymbol{W}_{down}^{(i)}\}^{4}_{i=1}$, respectively. The FFN can be rewritten as
\begin{equation}
\begin{aligned}
    \textit{FFN}(\boldsymbol{X})=\sum_{i=1}^4\textit{GELU}\left(\boldsymbol{X}\boldsymbol{W}_{up}^{(i)}\right)\boldsymbol{W}_{down}^{(i)}
\end{aligned}    
\end{equation}

In each layer, there are four $d\times d$ matrices in the MHSA block and eight $d\times d$ matrices in the FFN block. Supposing the number of layers in a ViT is $L$, we can stack all weights of the Transformer layers together as a single $12L\times d\times d$ tensor\footnote{We use superscript $j$ to represent that the matrix is in $j$-th layer.}
\begin{equation}
\begin{aligned}
    \boldsymbol{\mathcal{W}}=\big\{\{\boldsymbol{W}_q^j,& \boldsymbol{W}_k^j, \boldsymbol{W}_v^j, \boldsymbol{W}_o^j\}\cup \{\boldsymbol{W}_{up}^{j,(i)}\}^{4}_{i=1}\cup\\ &\{\boldsymbol{W}_{down}^{j,(i)}\}^{4}_{i=1} \big\}_{j=1}^{L}\in \mathbb{R}^{12L\times d\times d}
\end{aligned}    
\end{equation}
as shown in Fig~\ref{fig:tensorize}.

Note that the classification head, patch embedding, normalization, and all bias parameters are not taken into account in this tensorized format since they are irregular and few in number. For simplicity, we suppose the classification head is not tensorized, and others (patch embedding, normalization, and bias parameters) are frozen during fine-tuning. 

\subsection{Factor-Tuning: a Unified Perspective}
Let $\boldsymbol{\mathcal{W}}_0$ denote a tensorized pre-trained ViT. During fine-tuning, the ViT is updated to $\boldsymbol{\mathcal{W}}_{ft}$, and we use $\Delta\boldsymbol{\mathcal{W}}=\boldsymbol{\mathcal{W}}_{ft}-\boldsymbol{\mathcal{W}}_{0}$ to denote the {increment} of the ViT weight tensor. When fine-tuning, the gradient is calculated as\footnote{Tensor flattening notations are omitted for simplicity.}
\begin{equation}
g_{\boldsymbol{\mathcal{W}}} = \frac{\partial\mathcal{L}(\mathcal{D};\boldsymbol{\mathcal{W}})}{\partial\boldsymbol{\mathcal{W}}}
\label{eq1}
\end{equation}
where $\mathcal{D}$ is the training dataset and $\boldsymbol{\mathcal{W}}$ is initialized as $\boldsymbol{\mathcal{W}}_0$. We can rewrite this equation in another equivalent form
\begin{equation}
g_{\boldsymbol{\mathcal{W}}} = g_{\Delta\boldsymbol{\mathcal{W}}} = \frac{\partial\mathcal{L}(\mathcal{D};\boldsymbol{\mathcal{W}}_0+\Delta\boldsymbol{\mathcal{W}})}{\partial\Delta\boldsymbol{\mathcal{W}}}
\label{eq2}
\end{equation}
where $\Delta\boldsymbol{\mathcal{W}}$ is initialized as a zero tensor.

Traditional fine-tuning updates all parameters in a ViT, which means that we need to store at least a dense $\Delta\boldsymbol{\mathcal{W}}$ (for Eq (\ref{eq2})) or $\boldsymbol{\mathcal{W}}_{ft}$ (for Eq (\ref{eq1})) for each downstream task, resulting in a storage overhead of $\mathcal{O}(Ld^2)$ per task.

In the two modules of ViT, all the weight matrices multiply the inputs in a fully-connected way, implying the existence of redundancy. In the field of NLP, many studies have already found that the weight matrices in the transformer-based PLMs are redundant in rank~\cite{intra1,intra2,inter1,inter2}. The rank redundancies include \emph{intra-weight} redundancy, when each of the dense weight matrices can be decomposed (\emph{e.g.}, SVD) into and approximated by low-rank factors; and \emph{inter-weight} redundancy, when the model could work well with cross-layer shared weights (\emph{e.g.}, ALBERT~\cite{inter1}). Since the blocks of ViT and PLMs are highly similar, we infer that the weights of pre-trained ViT could be redundant in rank as well, suggesting that the rank of weight increments $\Delta\boldsymbol{\mathcal{W}}$ could also be redundant.

\begin{figure*}[t]
     \centering
         \includegraphics[width=0.80\textwidth]{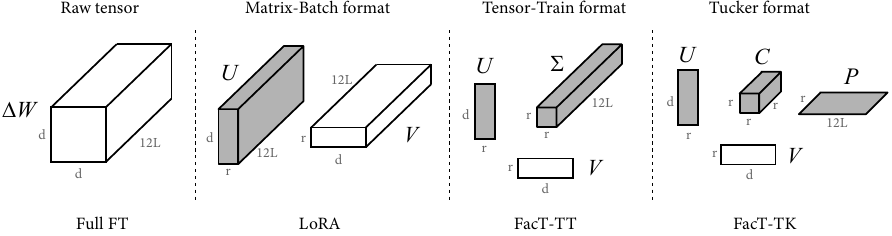}
         \caption{{Factor-Tuning with different tensor decomposition methods.} Full fine-tuning optimizes the raw tensor. LoRA can be regarded as optimizing factors of Matrix-Batch format. \texttt{\textbf{FacT-TT}} and \texttt{\textbf{FacT-TK}} optimize Tensor-Train and Tucker factors, respectively. Grey and white tensors/matrices indicate random and zero initialization, respectively.}
         \label{fig:formats}
\end{figure*}

\begin{figure}[t]
     \centering
         \includegraphics[width=0.4\textwidth]{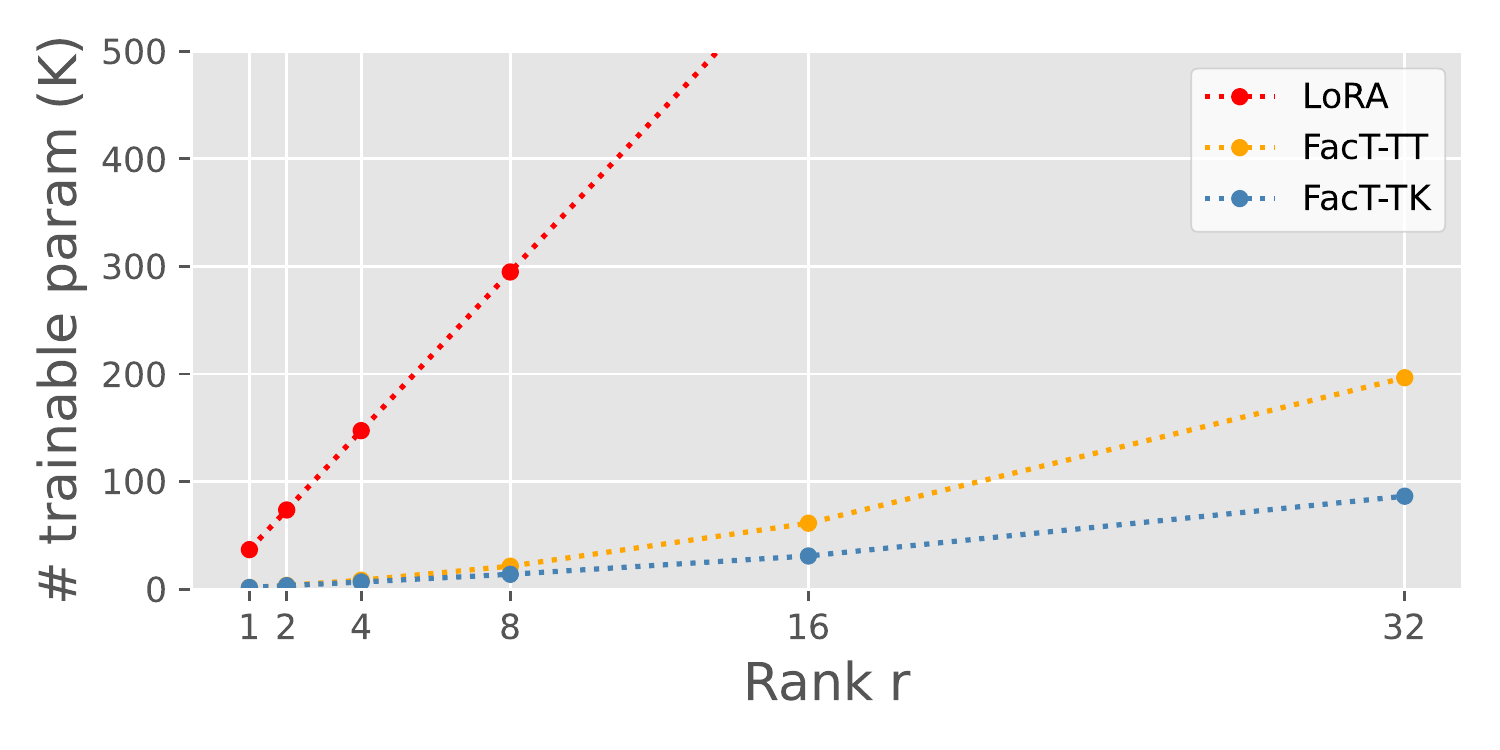}
         \caption{{Number of trainable parameters \emph{vs.} rank $r$.} With the same rank, \texttt{\textbf{FacT-TK}} uses fewer parameters than \texttt{\textbf{FacT-TT}}, and LoRA is significantly larger than \texttt{\textbf{FacT-TT}} and \texttt{\textbf{FacT-TK}}.}
         \label{fig:rank}
\end{figure}

Due to the redundancies of $\Delta\boldsymbol{\mathcal{W}}$, we can decompose $\Delta\boldsymbol{\mathcal{W}}$ into some factors to promote storage efficiency. We consider several well-known formats to decompose $\Delta\boldsymbol{\mathcal{W}}\in\mathbb{R}^{12L\times d\times d}$: Matrix-Batch format, Tensor-Train format~\cite{tt}, and Tucker format~\cite{tucker}, as illustrated in Fig~\ref{fig:formats}.

\subsubsection{Interpreting LoRA in our framework: Matrix-Batch}
Matrix-Batch format regards the first dimension of $\Delta\boldsymbol{\mathcal{W}}$ as the batch dimension and decomposes all the matrices of $d\times d$ in the batch individually. Formally, $\Delta\boldsymbol{\mathcal{W}}$ is decomposed into $\boldsymbol{U} \in \mathbb{R}^{12L\times d\times r}$ and $\boldsymbol{V} \in \mathbb{R}^{12L\times r\times d}$, where
\begin{equation}
\begin{aligned}
    \Delta\boldsymbol{\mathcal{W}}_{i,:,:}=s\cdot\boldsymbol{U}_{i,:,:}\boldsymbol{V}_{i,:,:}\qquad \forall i\in\{1,2,...,12L\}
\end{aligned}
\end{equation}
in which $s$ is a hyper-parameter for scaling. If we only consider $\boldsymbol{W}_q$ and $\boldsymbol{W}_v$ in tensorization, \emph{i.e.}, let $\boldsymbol{\mathcal{W}}=\big\{\boldsymbol{W}_q^j,\boldsymbol{W}_v^j\big\}_{j=1}^{L}\in \mathbb{R}^{2L\times d\times d}$ instead, \texttt{\textbf{FacT}} with Matrix-Batch format will become exactly LoRA. The size of LoRA's factors is $4Ldr\sim\mathcal{O}(Ldr)$, where $r<<d$.

However, since the weight matrices are discomposed individually in Matrix-Batch format, LoRA only reduces the intra-weight redundancy. We now introduce two other formats that take inter-weight redundancy into consideration.

\begin{table*}[t]

\centering
\setlength{\tabcolsep}{0.3pt}
\scalebox{0.95}{
\begin{tabular}{p{2.2cm}<{}p{0.9cm}<{\centering}|p{0.75cm}<{\centering}p{0.75cm}<{\centering}p{0.75cm}<{\centering}p{0.75cm}<{\centering}p{0.75cm}<{\centering}p{0.75cm}<{\centering}p{0.75cm}<{\centering}|p{0.75cm}<{\centering}p{0.75cm}<{\centering}p{0.75cm}<{\centering}p{0.75cm}<{\centering}|p{0.75cm}<{\centering}p{0.75cm}<{\centering}p{0.75cm}<{\centering}p{0.75cm}<{\centering}p{0.75cm}<{\centering}p{0.75cm}<{\centering}p{0.75cm}<{\centering}p{0.75cm}<{\centering}|p{0.75cm}<{\centering}}
\toprule[1.5pt]
\multicolumn{2}{c|}{}&\multicolumn{7}{c|}{\textbf{Natural}}&\multicolumn{4}{c|}{\textbf{Specialized}}&\multicolumn{8}{c|}{\textbf{Structured}}&\\
&\multicolumn{1}{c|}{\STAB{\rotatebox[origin=c]{90}{\# param (M)}}}
&\multicolumn{1}{c}{\STAB{\rotatebox[origin=c]{90}{Cifar100}}}
&\multicolumn{1}{c}{\STAB{\rotatebox[origin=c]{90}{Caltech101}}}
&\multicolumn{1}{c}{\STAB{\rotatebox[origin=c]{90}{DTD}}}
&\multicolumn{1}{c}{\STAB{\rotatebox[origin=c]{90}{Flower102}}}
&\multicolumn{1}{c}{\STAB{\rotatebox[origin=c]{90}{Pets}}}
&\multicolumn{1}{c}{\STAB{\rotatebox[origin=c]{90}{SVHN}}}
&\multicolumn{1}{c|}{\STAB{\rotatebox[origin=c]{90}{Sun397}}}
&\multicolumn{1}{c}{\STAB{\rotatebox[origin=c]{90}{Camelyon}}}
&\multicolumn{1}{c}{\STAB{\rotatebox[origin=c]{90}{EuroSAT}}}
&\multicolumn{1}{c}{\STAB{\rotatebox[origin=c]{90}{Resisc45}}}
&\multicolumn{1}{c|}{\STAB{\rotatebox[origin=c]{90}{Retinopathy}}}
&\multicolumn{1}{c}{\STAB{\rotatebox[origin=c]{90}{Clevr-Count}}}
&\multicolumn{1}{c}{\STAB{\rotatebox[origin=c]{90}{Clevr-Dist}}}
&\multicolumn{1}{c}{\STAB{\rotatebox[origin=c]{90}{DMLab}}}
&\multicolumn{1}{c}{\STAB{\rotatebox[origin=c]{90}{KITTI-Dist}}}
&\multicolumn{1}{c}{\STAB{\rotatebox[origin=c]{90}{dSpr-Loc}}}
&\multicolumn{1}{c}{\STAB{\rotatebox[origin=c]{90}{dSpr-Ori}}}
&\multicolumn{1}{c}{\STAB{\rotatebox[origin=c]{90}{sNORB-Azim}}}
&\multicolumn{1}{c|}{\STAB{\rotatebox[origin=c]{90}{sNORB-Ele}}}
&\multicolumn{1}{c}{\STAB{\rotatebox[origin=c]{90}{Average}}}\\
\specialrule{0em}{1pt}{1pt}
\hline
\specialrule{0em}{1pt}{1pt}
\multicolumn{22}{l}{\emph{Traditional Fine-Tuning}}\\
\hline
\specialrule{0em}{1pt}{1pt}
Full&85.8&68.9&87.7&64.3&97.2&86.9&87.4&38.8&79.7&95.7&84.2&73.9&56.3&58.6&41.7&65.5&57.5&46.7&25.7&29.1&68.9 \\
Linear&0&64.4&85.0&63.2&97.0&86.3&36.6&51.0&78.5&87.5&68.5&74.0&34.3&30.6&33.2&55.4&12.5&20.0&9.6&19.2&57.6\\
\hline
\specialrule{0em}{1pt}{1pt}
\multicolumn{22}{l}{\emph{PETL methods}}\\
\hline
\specialrule{0em}{1pt}{1pt}
BitFit&0.103&72.8&87.0&59.2&97.5&85.3&59.9&51.4&78.7&91.6&72.9&69.8&61.5&55.6&32.4&55.9&66.6&40.0&15.7&25.1&65.2\\
VPT-Shallow&\bf0.063&77.7&86.9&62.6&97.5&87.3&74.5&51.2&78.2&92.0&75.6&72.9&50.5&58.6&40.5&67.1&68.7&36.1&20.2&34.1&67.8\\
VPT-Deep&0.531&\bf78.8&90.8&65.8&98.0&88.3&78.1&49.6&81.8&\bf96.1&83.4&68.4&68.5&60.0&46.5&72.8&73.6&47.9&32.9&37.8&72.0 \\
Adapter&0.157&69.2&90.1&68.0&98.8&89.9&82.8&54.3&84.0&94.9&81.9&75.5&80.9&65.3&48.6&78.3&74.8&48.5&29.9&41.6&73.9 \\
AdaptFormer&0.157&70.8&91.2&\bf70.5&\bf99.1&\bf90.9&\bf86.6&\bf54.8&83.0&95.8&\bf84.4&\bf76.3&81.9&64.3&49.3&80.3&76.3&45.7&31.7&41.1&74.7 \\
LoRA&0.295&67.1&91.4&69.4&98.8&90.4&85.3&54.0&\bf84.9&95.3&\bf84.4&73.6&\bf82.9&\bf69.2&49.8&78.5&75.7&47.1&31.0&44.0&74.5
\\
NOAH&0.361&69.6&\bf92.7&70.2&\bf99.1&90.4&86.1&53.7&84.4&95.4&83.9&75.8&82.8&68.9&\bf49.9&\bf81.7&\bf81.8&\bf48.3&\bf32.8&\bf44.2&\bf75.5\\
\hline
\specialrule{0em}{1pt}{1pt}
\rowcolor{lightgray}\texttt{\textbf{FacT-TT}}$_{4}$&\bf0.008&69.4&88.5&70.6&98.8&90.0&83.3&53.7&83.9&95.1&81.5&75.4&78.2&\bf69.0&47.7&79.0&75.2&42.7&27.2&38.7&73.5\\
\rowcolor{lightgray}\texttt{\textbf{FacT-TT}}$_{\leq 16}$&0.037&\bf71.3&89.6&70.7&98.9&\bf91.0&87.8&\bf54.6&\bf85.2&95.5&83.4&\bf75.7&82.0&\bf69.0&\bf49.8&80.0&79.2&\bf48.4&\bf34.2&41.4&75.3\\
\rowcolor{lightgray}\texttt{\textbf{FacT-TK}}$_{8}$&0.014&70.3&88.7&69.8&99.0&90.4&84.2&53.5&82.8&95.6&82.8&\bf75.7&81.1&68.0&48.0&80.5&74.6&44.0&29.2&41.1&74.0\\
\rowcolor{lightgray}\texttt{\textbf{FacT-TK}}$_{\leq 32}$&0.069&70.6&\bf90.6&\bf70.8&\bf99.1&90.7&\bf88.6&54.1&84.8&\bf96.2&\bf84.5&\bf75.7&\bf82.6&68.2&\bf49.8&\bf80.7&\bf80.8&47.4&33.2&\bf43.0&\bf75.6\\
\bottomrule[1.5pt]
\end{tabular}
}

\caption{{Full results on the VTAB-1K benchmark}. ``\# params'' specifies the number of trainable parameters {in backbones}. Both average accuracy and \# params are averaged over group-wise average values. Our \texttt{\textbf{FacT-TK}}$_{\leq 32}$ outperforms all previous PETL methods while using significantly fewer parameters.}
\label{tab:vtab}
\end{table*}
\begin{figure*}[t]
     \centering
         \includegraphics[width=0.95\textwidth]{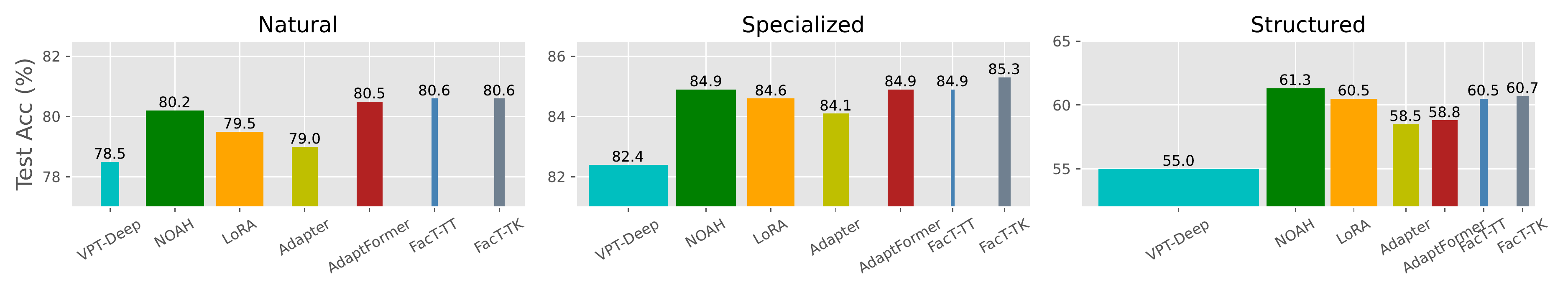}
         \caption{{Group-wise results on VTAB-1K.} The width of the bars is proportional to the number of trainable parameters.}
         \label{fig:group}
\end{figure*}

\subsubsection{Proposed Format I: Tensor-Train}
$\Delta\boldsymbol{\mathcal{W}}$ is decomposed into $\boldsymbol{U} \in \mathbb{R}^{d\times r_1}$, $\boldsymbol{V} \in \mathbb{R}^{d\times r_2}$, and $\mathcal{\boldsymbol{\Sigma}} \in \mathbb{R}^{12L \times r_1\times r_2}$, where
\begin{equation}
\Delta\boldsymbol{\mathcal{W}}=s\cdot\boldsymbol{\Sigma}\times_2\boldsymbol{U}^\intercal\times_3\boldsymbol{V}^\intercal
\end{equation}
in which $\times_i$ is mode-$i$ product, \emph{i.e.},
\begin{equation}
\begin{aligned}
\Delta\boldsymbol{\mathcal{W}}_{i,j,k}&=s\cdot\sum_{t_1=1}^{r_1}\sum_{t_2=1}^{r_2}\boldsymbol{\Sigma}_{i,t_1,t_2}\boldsymbol{U}_{j,t_1}\boldsymbol{V}_{k,t_2}\\& \forall i\in\{1,2,...,12L\}, \forall j,k\in\{1,2,...,d\}
\end{aligned}    
\end{equation}
For simplicity, we set $r=r_1=r_2<<d$. The size of factors is $2dr+12Lr^2\sim\mathcal{O}(dr+Lr^2)$. 
\subsubsection{Proposed Format II: Tucker}
$\Delta\boldsymbol{\mathcal{W}}$ is decomposed into $\boldsymbol{U} \in \mathbb{R}^{d\times r_2}$, $\boldsymbol{V} \in \mathbb{R}^{d\times r_3}$, $\boldsymbol{P} \in \mathbb{R}^{12L\times r_1}$, and $\boldsymbol{C} \in \mathbb{R}^{r_1 \times r_2\times r_3}$, where
\begin{equation}
\begin{aligned}
    &\Delta\boldsymbol{\mathcal{W}}=s\cdot\boldsymbol{C}\times_1\boldsymbol{P}^\intercal\times_2\boldsymbol{U}^\intercal\times_3\boldsymbol{V}^\intercal
\end{aligned}    
\end{equation}
\emph{i.e.},
\begin{equation}
\begin{aligned}
    \Delta\boldsymbol{\mathcal{W}}_{i,j,k}&=s\cdot\sum_{t_1=1}^{r_1}\sum_{t_2=1}^{r_2}\sum_{t_3=1}^{r_3}\boldsymbol{C}_{t_1,t_2,t_3}\boldsymbol{P}_{i,t_1}\boldsymbol{U}_{j,t_2}\boldsymbol{V}_{k,t_3}\\&\forall i\in\{1,2,...,12L\}, \forall j,k\in\{1,2,...,d\}
\end{aligned}    
\end{equation}
For simplicity, we set $r=r_1=r_2=r_3<<d$. The size of factors is $2dr+12Lr+r^3\sim\mathcal{O}(dr+Lr+r^3)$. 

Inspired by \citet{lora}, we adopt a decompose-then-train paradigm, \emph{i.e.}, decompose the $\Delta\boldsymbol{\mathcal{W}}$ before fine-tuning, and then update the factors end-to-end during fine-tuning. This paradigm benefits fine-tuning process in several ways. First, since $\Delta\boldsymbol{\mathcal{W}}$ is initially a zero tensor, we can use a pre-defined rule to initialize the factors directly instead of running the expensive decomposition algorithm. Second, decomposing a tensor means an uncontrollable loss of information. Because the factors are optimized via gradient descent, we can expect the most useful information to be retained when minimizing the training loss.

In each of the two formats, the factor $\boldsymbol{V}$ is zero-initialized and other factors are randomly initialized so that the $\Delta\boldsymbol{\mathcal{W}}$ is initially a zero tensor. After decomposition, we fine-tune the factors end-to-end. Taking the Tensor-Train format as an example, the gradient w.r.t. $\boldsymbol{U}$ is calculated as
\begin{equation}
\begin{aligned}
g_{\boldsymbol{U}} = \frac{\partial\mathcal{L}(\mathcal{D};\boldsymbol{\mathcal{W}}_0+\Delta\boldsymbol{\mathcal{W}})}{\partial\boldsymbol{{U}}}=s\cdot g_{\boldsymbol{\mathcal{W}}}\frac{\partial(\boldsymbol{\Sigma}\times_2\boldsymbol{U}^\intercal\times_3\boldsymbol{V}^\intercal)}{\partial\boldsymbol{U}}
\end{aligned}
\label{eq3}
\end{equation}
and the same for $\boldsymbol{V}$ and $\boldsymbol{\Sigma}$. Note that the role of hyper-parameter $s$ in Eq (\ref{eq3}) is to adjust the learning rate of factors.

After fine-tuning, we only need to store the lightweight factors for each task. We use \texttt{\textbf{FacT-TT}} and \texttt{\textbf{FacT-TK}} to denote \texttt{\textbf{FacT}} using Tensor-Train and Tucker formats, respectively. \texttt{\textbf{FacT-TT}} and \texttt{\textbf{FacT-TK}} reduce both intra- and inter-weight redundancies, so they can use fewer parameters to store task-specific information and are more storage-efficient, as shown in Fig~\ref{fig:rank}. The factors can be absorbed into $\boldsymbol{\mathcal{W}}_{ft}$ before inference, so \texttt{\textbf{FacT}} adds no extra computational cost or latency during the inference phase.

\section{Experiments}
\subsection{Transfer Learning on VTAB-1K Benchmark}
First of all, we evaluate our method on the basic transfer learning scenario -- fine-tuning the pre-trained models on various downstream tasks.

\subsubsection{Datasets} We use VTAB-1K benchmark~\cite{vtab} to evaluate the performance of our methods in terms of PETL. VTAB-1K consists of 19 different visual classification datasets, which can be divided into three groups: \textbf{Natural}, \textbf{Specialized}, and \textbf{Structured}.
Each dataset only contains 1,000 training samples. 
We report top-1 accuracy on test sets in all experiments. These datasets cover a large range of the possible domains where downstream tasks come from, and thus the effectiveness of PETL methods can be measured comprehensively.

\begin{figure*}[t]
     \centering
         \includegraphics[width=0.99\textwidth]{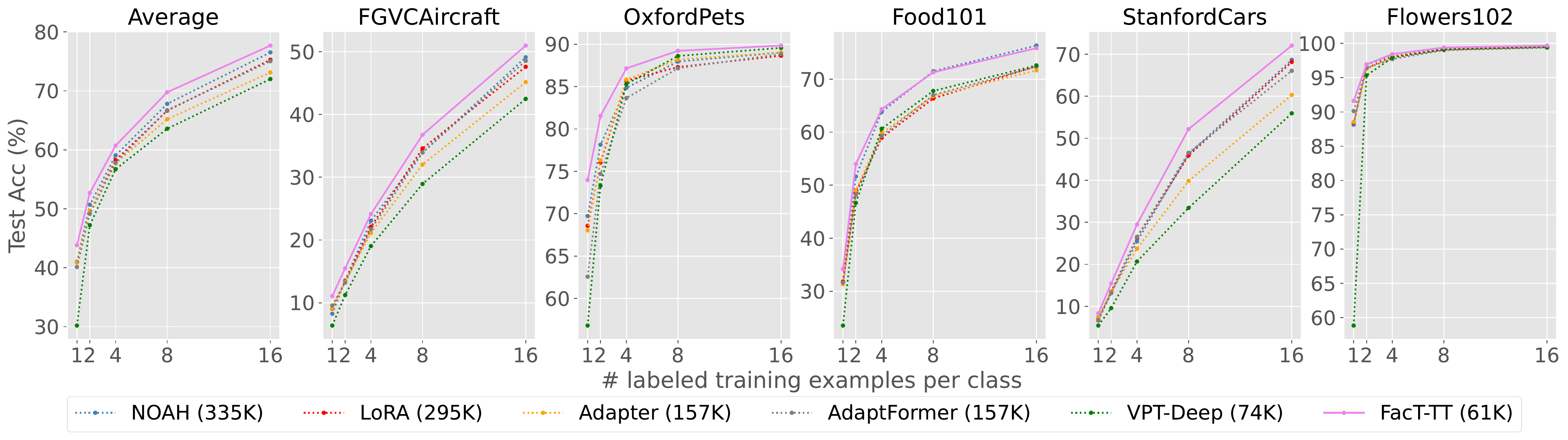}
         \caption{{Top-1 accuracy on fine-grained few-shot datasets.} The average numbers of trainable parameters in backbones are shown in parentheses. Our \texttt{\textbf{FacT-TT}} outperforms other baselines using the fewest trainable parameters.}
         \label{fig:fs}
\end{figure*}

\subsubsection{Compared Methods} We compare our methods to various competitive baselines, including \textbf{BitFit}~\cite{bitfit}, \textbf{VPT-Shallow}~\cite{vpt}, \textbf{VPT-Deep}~\cite{vpt}, \textbf{Adapter}~\cite{adapterp,compactor}, \textbf{AdapterFormer}~\cite{adaptformer}, \textbf{LoRA}~\cite{lora}, and current SOTA method \textbf{NOAH}~\cite{noah}. Following \citet{noah}, the hidden dimension $h$ of Adapter and AdaptFormer and the rank $r$ of LoRA are all set to 8. The prompt length $l$ of VPT follows the recipes in the original paper. We also report the results of two traditional transfer learning methods: \textbf{Full} fine-tuning, which updates all parameters on downstream tasks, and \textbf{Linear} probing, which learns a linear classification head on the pre-trained backbone.

For our methods, we report four settings: \texttt{\textbf{FacT-TT}}${_4}$ with $r=4$; \texttt{\textbf{FacT-TT}}$_{\leq 16}$ with $r$ searched from \{1, 2, 4, 8, 16\} for each task; \texttt{\textbf{FacT-TK}}$_{8}$ with $r=8$; and \texttt{\textbf{FacT-TK}}$_{\leq 32}$ with $r$ searched from \{2, 4, 8, 16, 32\}. Following \citet{noah}, we use AdamW optimizer with a learning rate of 1e-3 and batch size of 64 to train for 100 epochs. The hyper-parameter $s$ is roughly swept from \{0.01, 0.1, 1, 10, 100\}.

\subsubsection{Pre-trained Backbone} For all methods, we use a ViT-B/16~\cite{vit} pre-trained on supervised ImageNet-21K~\cite{imagenet} as the backbone. 

\subsubsection{Results} Experimental results are shown in Table~\ref{tab:vtab}, from which we can see that:

(1) {\texttt{\textbf{FacT-TT}} and \texttt{\textbf{FacT-TK}} have competitive results with respect to previous SOTA PETL methods while using much fewer trainable parameters.} \texttt{\textbf{FacT-TT}}$_{\leq 16}$ and \texttt{\textbf{FacT-TK}}$_{\leq 32}$ introduce only 37K and 69K trainable parameters, respectively. However, they outperform NOAH, the previous SOTA methods with 361K trainable parameters, on 11 out of 19 tasks. Moreover, \texttt{\textbf{FacT-TT}}$_{\leq 16}$ and \texttt{\textbf{FacT-TK}}$_{\leq 32}$ also achieve new SOTA results on 7 out of 19 tasks. It's worth noting that NOAH trains an additional large supernet for 500 epochs used for architecture search, and thus \texttt{\textbf{FacT}} is also superior to NOAH in terms of training efficiency.

(2) For our two methods, {neither \texttt{\textbf{FacT-TT}} nor \texttt{\textbf{FacT-TK}} shows an advantage over the other one.} Note that Tucker format can be regarded as further decomposing the $\boldsymbol{\Sigma}$ in Tensor-Train format into $\boldsymbol{P}$ and $\boldsymbol{C}$. Although Tucker format has a higher compression ratio, \texttt{\textbf{FacT-TK}} does not clearly outperform \texttt{\textbf{FacT-TT}}, implying that \texttt{\textbf{FacT-TT}} is sufficiently compact for adaptation and thus further compression does not lead to obvious improvement.

(3) Though sub-optimal to many baselines in terms of absolute performance, {\texttt{\textbf{FacT-TT}}$_{4}$ and \texttt{\textbf{FacT-TK}}$_{8}$ still provide non-trivial improvement under extreme storage constraints.} \texttt{\textbf{FacT-TT}}$_{4}$ and \texttt{\textbf{FacT-TK}}$_{8}$ use only 8K and 14K parameters (0.01\% and 0.02\% of the 85.8M ViT-B) to adapt the ViT backbones. In contrast, the parameters of the classification head will be even more than 8K as long as the task contains more than 10 classes. In other words, \texttt{\textbf{FacT-TT}}$_{4}$ and \texttt{\textbf{FacT-TK}}$_{8}$ use trainable parameters of the same magnitude as linear probing, while achieving performance better than full fine-tuning and VPT.

In Fig~\ref{fig:group}, we also report the group-wise results on VTAB-1K. \texttt{\textbf{FacT}} achieves SOTA results on Natural and Specialized, but underperforms NOAH on Structured. We still emphasize that our methods are much more lightweight, and thus more efficient than other baselines.

\subsection{Fine-Grained Few-Shot Learning}
Few-shot learning is a common scenario when the data of downstream tasks are hard to obtain, and there are only a few training samples for each task that can be utilized.
\subsubsection{Datasets}
To evaluate the capability of our method in the low-data regime, we conduct experiments on five fine-grained datasets in few-shot settings. The five datasets include \textbf{FGVC-Aircraft}~\cite{aircraft}, \textbf{Oxford-Pets}~\cite{pets}, \textbf{Food-101}~\cite{food}, \textbf{Stanford Cars}~\cite{car}, and \textbf{Oxford-Flowers102}~\cite{flower}, which contains fine-grained classes from five categories: aircraft, pets, food, cars, and flowers. Following previous work~\cite{noah}, we evaluate in \{1, 2, 4, 8, 16\}-shot settings.

\subsubsection{Compared Methods}
We compare our method with five baselines that perform the best on VTAB-1K: VPT-Deep, Adapter, AdaptFormer, LoRA, and NOAH. The hyper-parameter $h$, $r$, and $l$ of the baselines are all set to 8. As for our method, we report the results of \texttt{\textbf{FacT-TT}}$_{16}$, \emph{i.e.}, \texttt{\textbf{FacT-TT}} with a fixed $r=16$. Other settings are the same as on VTAB-1K. All results are averaged over three runs with different random seeds.

\subsubsection{Results}
As the results shown in Fig~\ref{fig:fs}, we can find that:

(1) Though using the fewest trainable parameters, \texttt{\textbf{FacT-TT}} still achieves SOTA results on average. Note that all these datasets can be categorized into the Natural group, so this observation is in line with what we have found on VTAB-1K that \texttt{\textbf{FacT-TT}}  is better at Natural tasks.

(2) \texttt{\textbf{FacT-TT}} performs the best across all settings on four out of five datasets except for Food-101, where \texttt{\textbf{FacT-TT}} slightly underperforms NOAH in 8-shot and 16-shot settings.

These observations confirm the capability and efficiency of our method in the low-data regime. Again, it verifies the effectiveness of reducing intra- and inter-rank redundancies of weight increments for PETL. 


\subsection{\texttt{\textbf{FacT}} for Hierarchical Transformers}
After the original ViT was proposed, it was found that ViT lacks visual inductive bias which limits its performance, especially on dense prediction tasks. Subsequently, a series of studies improved ViT by introducing hierarchical structures~\cite{swin,pvt,cvt}, among which Swin Transformer~\cite{swin} is a widely used and representative design. Therefore, we also extend \texttt{\textbf{FacT}} to Swin Transformer.

A challenge when applying tensorization to such hierarchical structures is that the hidden dimension $d$ is different across layers. However, these models usually partition the layers into several stages, where the hidden dimension is consistent within each stage. Therefore, we propose a \emph{partitioned tensorization} strategy for these models, which individually tensorizes each stage into a single tensor.

Taking \textbf{Swin-B} as an instance, its four stages consist \{2, 2, 18, 2\} layers with hidden dimensions of \{128, 256, 512, 1024\}, respectively. Therefore, we can tensorize them into four tensors of size \{24$\times$128$\times$128, 24$\times$256$\times$256, 216$\times$512$\times$512, 24$\times$1024$\times$1024\} following the steps described in Section 3.1. These tensors are then decomposed individually. 

We report the results of \texttt{\textbf{FacT-TT}}$_{16}$ on VTAB-1K in Table~\ref{tab:fa3}, using Swin-B pre-trained on supervised ImageNet-21K as the backbone. We compare our method with baselines that can also be employed on Swin: Full, Linear, BitFit, and VPT. We can see that our \texttt{\textbf{FacT-TT}} outperforms other PETL methods by a large margin. Although partitioned tensorization weakens the efficiency of \texttt{\textbf{FacT}} to some extent since we have to store a set of factors for each stage, \texttt{\textbf{FacT-TT}} still uses fewer parameters than VPT-Deep and BitFit. These results demonstrate that \texttt{\textbf{FacT}} can also be applied to Swin Transformer while keeping its advantages. Since \texttt{\textbf{FacT}} is an architecture-agnostic framework, it can be extended to various models (\emph{e.g.}, ConvNets~\cite{convnext} and MLPs~\cite{mlp,asmlp}) and tasks (\emph{e.g.}, NLP and multimodal tasks), as long as the models can be tensorized appropriately. 
\begin{table}[t]
\centering
\scalebox{0.95}{
\begin{tabular}{p{1.9cm}<{\centering}p{1.8cm}<{\centering}p{0.6cm}<{\centering}p{0.6cm}<{\centering}p{0.6cm}<{\centering}p{0.6cm}<{\centering}}
\toprule[1.5pt]
\specialrule{0em}{1pt}{1pt}
Method&\# param (M)&Avg.&Nat.&Spe.&Str.\\\hline
\specialrule{0em}{1pt}{1pt}

Full&86.7&75.0&79.2&86.2&59.7\\\specialrule{0em}{1pt}{1pt}
Linear&0&62.6&73.5&80.8&33.5\\\specialrule{0em}{1pt}{1pt}
BitFit&0.201&65.6&74.2&80.1&42.4\\\specialrule{0em}{1pt}{1pt}

VPT-Shallow&0.003&66.7&79.9&82.5&37.8\\\specialrule{0em}{1pt}{1pt}
VPT-Deep&0.162&71.6&76.8&84.5&53.4\\\specialrule{0em}{1pt}{1pt}
\rowcolor{lightgray}\texttt{\textbf{FacT-TT}}$_{16}$&0.135&\bf77.4&\bf83.1&\bf86.9&\bf62.1\\
\bottomrule[1.5pt]
\end{tabular}
}
\caption{{Results on VTAB-1K with Swin-B as backbone. }Avg./Nat./Spe./Str.: Average/Natural/Specialized/Structured results. \# params: \# of trainable parameters {in backbones.}}
\label{tab:fa3}
\end{table}

\begin{figure}[t]
     \centering
         \includegraphics[width=0.475\textwidth]{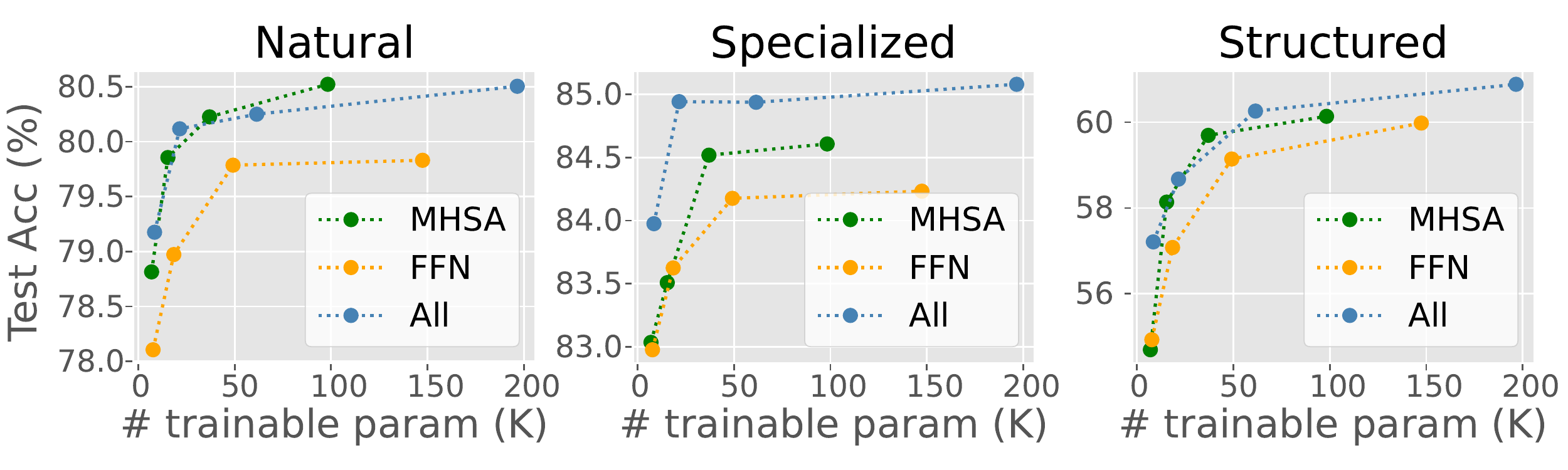}
         \caption{{Results on VTAB-1K across different ranks and tensorization strategies.} All: the original tensorization method that tensorizes both MHSA and FFN; MHSA/FFN: tensorize MHSA/FFN blocks only. We report results with $r\in$ \{4, 8, 16, 32\} for each setting.}
         \label{fig:fa}
\end{figure}

\subsection{Ablation Analyses}
We ablate the tensorization-decomposition framework on VTAB-1K benchmark to show the effect of tensorization strategy and decomposition rank.

The default tensorization method tensorizes both MHSA and FFN blocks, resulting in a $12L\times d\times d$ tensor. We here consider two ablated strategies: tensorizing only MHSA or FFN blocks, \emph{i.e.},
\begin{equation}
    \boldsymbol{\mathcal{W}}=\big\{\boldsymbol{W}_q^j, \boldsymbol{W}_k^j, \boldsymbol{W}_v^j, \boldsymbol{W}_o^j\big\}_{j=1}^{L} \in \mathbb{R}^{4L\times d\times d}
\end{equation}
or
\begin{equation}
    \boldsymbol{\mathcal{W}}=\big\{\{\boldsymbol{W}_{up}^{j,(i)}\}^{4}_{i=1}\cup \{\boldsymbol{W}_{down}^{j,(i)}\}^{4}_{i=1} \big\}_{j=1}^{L} \in \mathbb{R}^{8L\times d\times d}
\end{equation}
respectively. The blocks that are not tensorized keep frozen during fine-tuning. As for the decomposition part, we use Tensor-Train format with rank $r\in$  \{4, 8, 16, 32\}. The results are shown in Fig \ref{fig:fa}.

On all three groups, tensorizing MHSA is better than tensorizing FFN overall, suggesting that MHSA blocks play a more important role than FFN in downstream transfer tasks. Tensorizing all blocks is better than the other two strategies on Specialized and Structured. We also find that only tensorizing MHSA is slightly superior to tensorizing all on Natural when $r$ is larger, indicating that it is feasible to further develop the potential of \texttt{\textbf{FacT}} by searching the tensorization strategy for different datasets. Besides, we find that as rank $r$ increases, the average performance is accordingly improved across the three tensorization settings. However, since the size of factor $\boldsymbol{\Sigma}$ increases with the square of $r$, \texttt{\textbf{FacT-TT}} is no longer efficient when $r$ becomes too large. But this limitation will not have a significant negative impact because the performance is almost saturated when $r\geq16$ and thus a large $r$ is not necessary.

\section{Conclusion}
In this paper, we propose \texttt{\textbf{FacT}}, a tensorization-decomposition framework for PETL on ViT. Under this framework, we present an approach to making ViT tensorized, and employ two tensor decomposition methods to factorize its increment. By updating and storing the factors only, our methods reduce both intra- and inter-weight redundancies of weight increments and thus are much more efficient. \texttt{\textbf{FacT}} achieves competitive results on VTAB-1K benchmark with a significantly reduced number of parameters and outperforms all PETL baselines on few-shot learning. Our work demonstrates that the storage efficiency of PETL has not been fully exploited yet, and \texttt{\textbf{FacT}} provides a promising framework for future work. 

\bibliography{aaai23}
\iftrue
\onecolumn
\section*{Appendix}
\setcounter{section}{0}

\renewcommand\thesection{\Alph{section}}
\section{Datasets}
See Table~\ref{tab:dataset}. The 1,000 training samples of VTAB-1K are further divided into a training set (800 samples) and a validation set (200 samples) in hyper-parameter tuning. The final results are produced by training on all 1,000 training samples and testing on the test set.
\begin{table*}[h]

\centering
\scalebox{0.9}{
\begin{tabular}{clcccc}
\toprule[1.5pt]
                          & Dataset              & \# Classes & Train                    & Val   & Test            \\ \midrule
                          \multicolumn{6}{c}{VTAB-1K~\cite{vtab}}\\\midrule
\multirow{7}{*}{Natural} & CIFAR100~\cite{krizhevsky2009learning}             & 100       & \multirow{7}{*}{800/1,000}                 & \multirow{7}{*}{200}   & 10,000           \\
                          & Caltech101~\cite{fei2004learning}           & 102       &                 &    & 6,084            \\
                          & DTD~\cite{cimpoi14describing}                 & 47        &                 &    & 1,880            \\
                          & Oxford-Flowers102~\cite{flower}    & 102       &                 &    & 6,149            \\
                          & Oxford-Pets~\cite{pets}          & 37        &                 &    & 3,669     \\
                          & SVHN~\cite{netzer2011reading}                 & 10        &                 &    & 26,032            \\
                          & Sun397~\cite{xiao2010sun}               & 397       &                 &    & 21,750           \\
\midrule\multirow{4}{*}{Specialized}                          & Patch Camelyon~\cite{Veeling2018qh}       & 2         &      \multirow{4}{*}{800/1,000}                 & \multirow{4}{*}{200}                & 32,768           \\
                          & EuroSAT~\cite{helber2019eurosat}              & 10        &                 &    & 5,400            \\
                          & Resisc45~\cite{cheng2017remote}             & 45        &                 &    & 6,300            \\
                          & Retinopathy~\cite{kaggle2015retinopathy}         & 5         &                 &    & 42,670           \\
 \midrule\multirow{8}{*}{Structured}                         & Clevr/count~\cite{johnson2017clevr}          & 8         &      \multirow{8}{*}{800/1,000}                 & \multirow{8}{*}{200}                  & 15,000       \\
                          & Clevr/distance~\cite{johnson2017clevr}       & 6         &                 &    & 15,000     \\
                          & DMLab~\cite{beattie2016deepmind}                & 6         &                 &    & 22,735           \\
                          & KITTI-Dist~\cite{geiger2013vision}           & 4         &                 &    & 711   \\
                          & dSprites/location~\cite{matthey2017dsprites}    & 16        &                 &    & 73,728           \\
                          & dSprites/orientation~\cite{matthey2017dsprites} & 16        &                 &    & 73,728           \\
                          & SmallNORB/azimuth~\cite{lecun2004learning}    & 18        &                 &    & 12,150     \\
                          & SmallNORB/elevation~\cite{lecun2004learning}  & 18         &                 &    & 12,150     \\ \midrule
                           \multicolumn{6}{c}{Few-shot learning}\\\midrule
 & Food-101~\cite{food}             & 101       & \multirow{5}{*}{1/2/4/8/16 per class} & 20,200 & 30,300            \\
                          & Stanford Cars~\cite{car}        & 196       &  & 1,635  & 8,041       \\
                          & Oxford-Flowers102~\cite{flower}    & 102       &  & 1,633  & 2,463             \\
                          & FGVC-Aircraft~\cite{fgvc}        & 100       &  & 3,333  & 3,333       \\
                          & Oxford-Pets~\cite{pets}          & 37        &  & 736   & 3,669    \\
\bottomrule[1.5pt]
\end{tabular}}
\caption{\textbf{Statistics of used datasets.}
}
\label{tab:dataset}
\end{table*}

\section{Experimental Details}
\subsection{Pre-trained Backbones}
See Table~\ref{tab:pt}.

\begin{table*}[h]

\centering
\begin{minipage}{\linewidth}
\centering

\scalebox{1}{

\begin{tabular}{lccc}
\toprule[1.5pt]
Model&Pre-training Dataset&Size (M)&Pre-trained Weights
\\\midrule
ViT-B/16~\cite{vit}&ImageNet-21K&85.8&\footnote{https://storage.googleapis.com/vit\_models/imagenet21K/ViT-B\_16.npz}{checkpoint}
\\
Swin-B~\cite{swin}&ImageNet-21K&86.7&\footnote{https://github.com/SwinTransformer/storage/releases/download/v1.0.0/swin\_base\_patch4\_window7\_224\_22k.pth}{checkpoint}
\\
 \bottomrule[1.5pt]
\end{tabular}
}
\end{minipage}

\caption{\textbf{Pre-trained backbones.}
}
\label{tab:pt}
\end{table*}

\subsection{Code Implementation}
We use \footnote{https://pytorch.org/}{\emph{PyTorch}} and \footnote{https://rwightman.github.io/pytorch-image-models/}{\emph{timm}} to implement all experiments on NVIDIA RTX 3090 GPUs.

\subsection{Data Augmentation}
\subsubsection{VTAB-1K} Following \citet{noah}, we resize the images to $224\times224$, and then \textbf{normalize them with ImageNet's mean and standard deviation}.

{We notice that some methods like VPT~\cite{vpt} and concurrent SSF~\cite{ssf} do not normalize the input images of VTAB-1K. Such a setting could benefit downstream tasks by closing the gap between the inputs of pre-training and fine-tuning, since the inputs are also not normalized in the pre-training process of ViT-B. We find that the average accuracy on VTAB-1K of PETL methods (including ours) could gain about 1-2\% by using non-normalized inputs. However, since we reuse some results reported by NOAH, we still adopt the sub-optimal pipeline (\emph{i.e.}, normalized inputs as in NOAH) for a fair comparison with the state-of-the-art.}
\subsubsection{Few-shot learning} Following \citet{noah}, for training samples, we use color-jitter and RandAugmentation; for validation/test samples, we resize them to $256\times256$, crop them to $224\times224$  at the center, and then normalize them with ImageNet's mean and standard deviation.

\subsection{Hyper-parameters}
$s$ is searched from \{0.01, 0.1, 1, 10, 100\}. See Table~\ref{tab:hyper} for other hyper-parameters. We basically follow the hyper-parameters used by \citet{noah}.
\begin{table*}[h]

\centering
\scalebox{0.9}{
\begin{tabular}{cccccccc}
\toprule[1.5pt]
 &optimizer&batch size&learning rate&weight decay&\# epochs&lr decay&\# warm-up epochs\\ \midrule
 VTAB-1K&AdamW&64&1e-3&1e-4&100&cosine&10\\
 Few-shot learning&AdamW&64&5e-3&1e-4&100&cosine&10\\
 \bottomrule[1.5pt]
\end{tabular}}
\caption{\textbf{Hyper-parameters.}
}
\label{tab:hyper}
\end{table*}

\subsection{Reproducibility of Baselines}
The results of baselines can be reproduced by the official codebases of VPT~\cite{vpt} and NOAH~\cite{noah}.

We notice that it is reported by \citet{vpt} that Adapter significantly underperforms VPT. This is because their implementation uses zero-initialization for weights of both the two FC layers in Adapter, which blocks the back-propagation of gradients. Therefore, we follow \citet{noah} using Xavier-initialization for weights and zero-initialization for biases, and report results that Adapter outperforms VPT. Similar observations are also reported in NLP literature~\cite{adapter,towards}.
\twocolumn
\fi

\end{document}